# An ascription-based approach to Speech Acts

**Mark Lee and Yorick Wilks**
Department of Computer Science
University of Sheffield
Regent Court, 211 Portobello Street
Sheffield S1 4DP, UK
*M.Lee@dcs.shef.ac.uk*
*Y.Wilks@dcs.shef.ac.uk*

## Abstract:

The two principal areas of natural language processing research in pragmatics are belief modelling and speech act processing. Belief modelling is the development of techniques to represent the mental attitudes of a dialogue participant. The latter approach, speech act processing, based on speech act theory, involves viewing dialogue in planning terms. Utterances in a dialogue are modelled as steps in a plan where understanding an utterance involves deriving the complete plan a speaker is attempting to achieve. However, previous speech act based approaches have been limited by a reliance upon relatively simplistic belief modelling techniques and their relationship to planning and plan recognition. In particular, such techniques assume precomputed nested belief structures. In this paper, we will present an approach to speech act processing based on novel belief modelling techniques where nested beliefs are propagated on demand.

## 1. Introduction

The use of simplistic belief models has accompanied complex accounts of speech acts where highly nested belief sets accompany any speech act. We believe that by utilising a more sophisticated view of mental attitudes, a simpler and more elegant theory of speech acts can be constructed. Also, as previous work has pointed out (Wilks et al, 1991) past models have failed to differentiate explicitly between the speaker's and hearer's belief sets. Such a failure causes problems in dealing with misconceptions and badly formed plans (Pollack, 1990).

This paper augments ViewGen, a computer program originally developed by Ballim and Wilks (1991) to model the beliefs and meta-beliefs of a system using nested belief structures. ViewGen is able to reason about its own and other agent's beliefs using belief ascription and inference techniques. The current version of ViewGen is implemented in Quintus Prolog.

The structure of this paper is as follows: in Section 2, we review and discuss previous speech act approaches and their representation of mental attitudes. We argue that precomputed highly nested belief structures aren't necessary. In Section 3, we describe how ViewGen represents mental attitudes and computes nested structures by a process of ascription and in Section 4, show how such techniques can be used to represent speech acts for use in planning and plan recognition. Finally, in Section 5, we discuss some implications and future directions of our work

## 2. Speech acts and mental attitudes

It is clear that any understanding of an utterance must involve reference to the attitudes of the speaker. For example, the full understanding of the utterance "Do you know where Thomas is?" depends upon whether the speaker already knows where Thomas is and whether he or she believes the hearer knows.

Speech act based AI approaches normally make reference to mental attitudes and often provide links between the surface form of the utterance and the mental attitudes of both the speaker and hearer. For example, Appelt (1985) describes a system which generates discourse from an intensional logic representation of a set of beliefs. However, as pointed out by Pollack (1990), they have typically used relatively simple models of such attitudes. In particular, previous approaches have lacked any way to model the propagation of belief within the system itself and instead have made use of precomputed and fixed nestings of mental attitudes.

One widely used concept in speech act accounts is mutual belief. Following work in philosophy by Lewis (1969), Clark and Marshall (1981) introduced the notion of mutual belief to account for hearer attitudes. A proposition P is a mutual belief if shared by two agents A and B such that:

<div align="center">
A believes P
B believes P
A believes B believes P
B believes A believes P
etc., ad infinitum
</div>

There cannot be a logical limit to the number of levels of regression since, as Schiffer (1972) argued, for any level of nested belief, a dialogue example can be constructed which requires an additional level of belief nesting. Because of this potentially infinite regression, it has proven difficult to use an axiomatic definition of mutual belief based in terms of simple belief in computational implementations. Alternative approaches have either avoided defining axioms for mutual belief, e.g. Taylor and Whitehill (1981) or defined it as a primitive operator without reference to simple beliefs, e.g. Cohen and Levesque (1985).

Despite such work, it appears that the mutual belief hypothesis, i.e. that agents compute potentially infinite nestings of belief in comprehension, appears to be too strong a hypothesis to be realistic. It is impossible that agents perform this kind of potentially infinite nesting during real dialogue and no clear constraint can be given on how many iterations would be necessary in a real dialogue situation. Though examples can be artificially created which require n levels of nesting for large n, during a study of dialogue corpora, Lee (1994) found no need for highly nested belief models. In fact, it appears that no dialogue exchange required more than a two level belief nesting. Also, mistakes in assuming what was common to both agents in a dialogue occurred but were quickly repaired through the use of corrections and repetitions and other dialogue control acts. Similar results have been reported by Taylor and Carletta (1994) in analysing the HCRC Map Task corpus.

Rather than compute nested beliefs to some fixed level during comprehension. It is far more plausible that agents compute nested representations on so that highly nested belief representations are only constructed if required in the dialogue. This is the basic principle behind ViewGen.

## 3. The ViewGen system

ViewGen is a nested attitude model which constructs intensional environments to model the attitudes of other agents. Previous work on ViewGen has been concerned with only modelling belief attitudes (Wilks, Barnden and Ballim, 1991). We have extended ViewGen to model and represent, in addition, goals and intentions. In this section, we briefly describe ViewGen's operation.

### 3.1 ViewGen representations of mental attitudes

ViewGen assumes that each agent in a dialogue has a belief environment which includes attitudes about what other agents believe, want, and intend. Such attitudes are represented in a nested structure. Each nesting is an environment which contains propositions which may be grouped by a particular topic or stereotype. The particular topic is given on the top left corner of the environment while the holder of a belief is given at the bottom of the environment.

ViewGen represents all attitudes in environments with the attitude type labelled on the far right bottom of the box. Though different attitude types are separated by environments, they can be nested so that agents can have beliefs, goals, and intentions about these attitudes. For example, suppose the System believes that John intends to buy a car, but wants to convince him otherwise by getting him to believe correctly that the car is a wreck. This is illustrated in Figure 1.

In ViewGen, different attitudes have different types. Beliefs and goals refer to propositions which the agent either believes is true or wants to be true at some point in the future. Intentions, however, are represented as connected planning actions which represent the plans the agent intends to pursue to achieve his or her goals.

### 3.2 Ascription of attitudes

As noted above, ViewGen avoids using a concept of shared or mutual beliefs. Rather, ViewGen attributes beliefs, goals and intentions to other agents as required. This process is termed *ascription*. There are two methods of ascription: default ascription and stereotypical ascription. Each method is briefly described below.

### 3.2.1 Default Ascription

Default ascription applies to common beliefs. Most beliefs in any set held by an agent are common beliefs about the world, and can be assumed to be common to any other rational agent unless marked otherwise. For example, an agent may believe that the world is round and therefore, without any evidence, guess that any other agent probably shares this belief. To model this, ViewGen uses a *default ascription rule* i.e.

**Default Ascription rule:**
*Given a System belief, ascribe it to any other agent as required, unless there is contrary evidence.*

Such a rule results in beliefs being pushed from outer belief environments to inner belief environments. For example, Figure 2 illustrates ViewGen assuming that John shares its belief that the world is round.

Evidence against ascription is normally an explicit belief

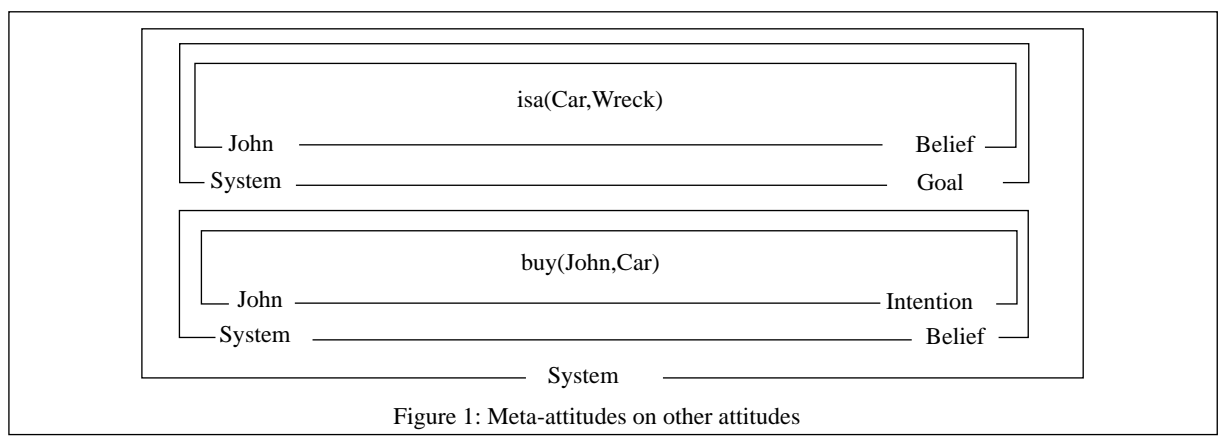

Figure 1: Meta-attitudes on other attitudes

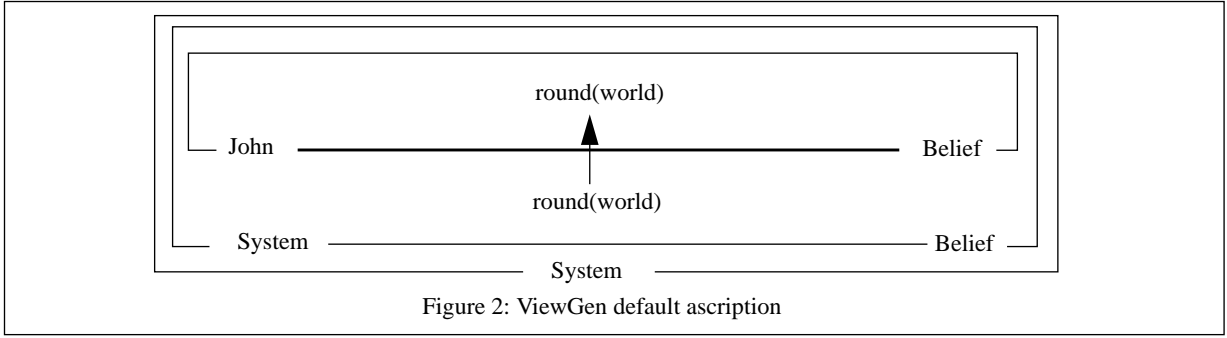
Figure 2: ViewGen default ascription

that an another agent believes the opposite of the ascribed belief. For example, an agent might already be believed by ViewGen to believe the world is flat and thus block any ascription of ViewGen's belief. It is important for ViewGen to reason about other agent's beliefs about other agents. For example, it is plausible that an agent who believes the world is round may well also believe by default that other agents believe the same.

Unlike beliefs, the assumption that other agents share similar goals and intentions cannot be made by default. Goals and intentions are more dynamic than beliefs in that an agent will try to achieve goals and carry out intentions in the future which once achieved are dropped from the agent's attitude set. Also, goals and intentions are often highly stereotypical. Therefore, a default rule of ascription cannot be applied to such attitudes. However, a combination of stereotypical ascription and plan recognition can be used to provide sensible ascriptions of goals and intentions. Stereotypical ascription is discussed next while plan recognition is discussed in 4.3.

### 3.2.2 Stereotypical Ascription

A stereotype is a collection of attitudes which are generally applicable to a particular class of agent. For example, Doctors tend to have expert medical knowledge and have goals to diagnose diseases and cure patients. To model this ViewGen uses a stereotypical ascription rule:

**Stereotypical Ascription rule:**
*Given a System stereotypical belief, ascribe it to any other agent to which the stereotype applies as required, unless there is contrary evidence.*

In ViewGen, stereotypes consist of sets of attitudes which an agent who fits a particular stereotype might typically hold. Such stereotypical beliefs can be ascribed to an agent by default - i.e. unless there is explicit evidence that the agent holds a contrary belief. For example, in Figure 3, the System has a stereotypical set of beliefs for doctors and, since it believes John is a doctor, ascribes these to John.

## 4. Ascription based Speech act representation

In this section, we will outline our theory of speech acts. In 4.1, we outline a list of features which we believe any theory should possess and in 4.2 we describe a theory based on belief ascription.

### 4.1 Desideratum for a theory of speech acts

We believe that a theory of speech acts should have at least the following features:

1, *The theory should be solipsist*
The notion of mutual knowledge was introduced to provide a realistic account of the effects of a speech act on a hearer. However, as has argued above and elsewhere (Ballim and Wilks, 1991), mutual belief is too strong a notion to be used. Instead, a theory of speech acts should be solipsistic in that it refers solely to finite belief representations of either the speaker or hearer of the dialogue act.

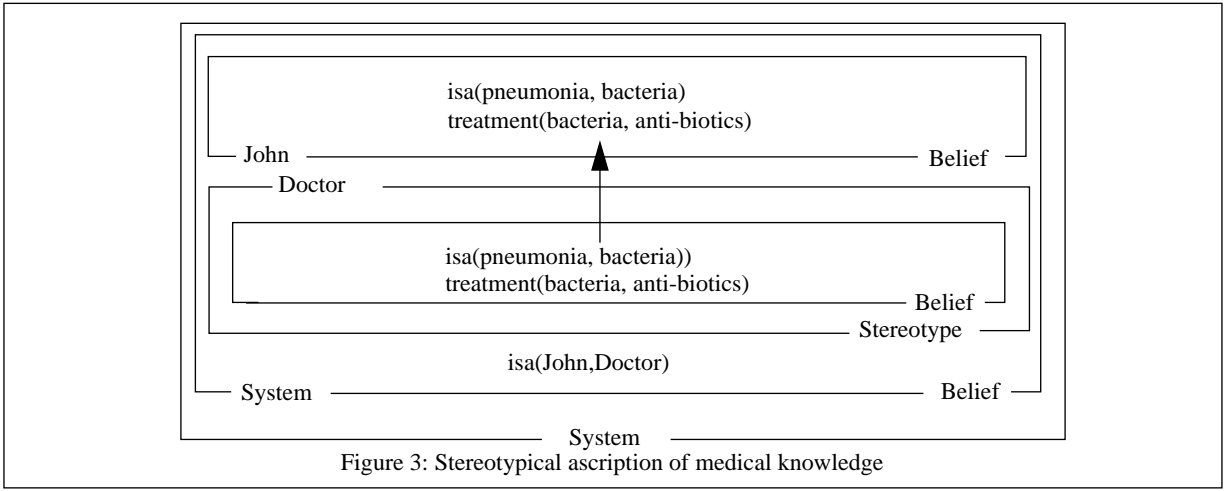
Figure 3: Stereotypical ascription of medical knowledge

2, *The theory must provide separate interpretations for the speaker and hearer*

A theory must, however, take into account the attitudes of both the speaker and hearer by allowing the separate derivation of the effects of a speech act from the speaker's and the hearer's points of view.

3, *Speech acts should be minimalistic*

Any theory should assume only the minimal conditions for any utterance to be successful. This means avoiding the ascription of precomputed attitude nestings beyond the essential conditions of each act for the act to be achieved.

4, *Speech acts should be extendable*

Despite assuming only the minimal conditions and effects for any speech act, they should in principle be extendable to deal with problematic examples involving high degrees of belief nesting proposed by work in philosophy.

5, *The theory must provide a means to derive generalised effects from each acts conditions*

As argued by Searle (1969), any classification of speech acts must be based on the conditions of each act and not its effects. However, we also want a principled way to derive the conventional effects of any act from its conditions. This is necessary so that we can then provide a clear distinction between an act's conventional illocutionary effect and its context-specific perlocutionary effect.

We believe that our account of speech acts satisfies the above criteria. In the next two sections we will outline how we represent speech acts in terms of belief ascription and how we use these in planning and plan recognition.

## 4.2 An ascription based theory of speech acts

We represent 20 different speech acts types in four classes: questions, answers, requests and inform acts. This set is partially based on Bunt's taxonomy of 24 speech acts (1989). While not claiming that such a set of acts is complete, we have found it sufficient for the dialogue corpora we have analysed. Every act is classified with respect to its preconditions which are the mental attitudes a speaker must adopt to felicitously perform the speech act. Acts are ordered by specificity: more specific speech acts inherit or strengthen the preconditions of more general ones. For example, an inform act requires that the speaker believes the proposition in question and has a goal that the hearer also believes the proposition, i.e.:

Inform(Speaker,Hearer,Proposition)
Preconditions: believe(Speaker,Proposition)
goal(Speaker,believe(Hearer,Proposition))

A correction act is a more specific type of informing and, therefore, inherits the preconditions of informing plus the condition that the speaker believes that the hearer believes the opposition of the proposition, i.e.:

Correction(Speaker,Hearer,Proposition)
Preconditions: believe(Speaker,Proposition)
goal(Speaker,believe(Hearer,Proposition))
believe(Speaker,believe(Hearer,
not(Proposition)))

Rather than specify individual effects for each dialogue act, we provide separate update rules based on belief ascription. Our update rule from the speaker's point of view is:

Update on the Speaker's belief set
*For every condition C in dialogue act performed:*
*default_ascribe(Speaker, Hearer, believe(C))*

That is, for every condition in the speech act, the speaker must ascribe a belief to the hearer that the condition is satisfied. For example, Figure 4 shows the conditions for an inform act: the speaker believes the proposition to be communicated and wants the hearer to believe it too. To achieve this goal, the speaker intends to use an inform speech act. After performing the inform act, the speaker can ascribe to the hearer the belief that each of the preconditions were met i.e. the speaker believes that the hearer believes the speaker believes the proposition and has the goal of getting the hearer to believe it too. The effects of the inform act on the speaker's attitude set are shown in Figure 5. Note that after the inform act is performed, the intention to perform it is dropped. However, the speaker's goal of getting the hearer to believe the proposition remains. This is because we assume only the minimal conditions for the act to be successful i.e. if the speaker can successfully ascribe each speech act precondition to the hearer. For the hearer to believe the proposition, he or she has to perform a mental

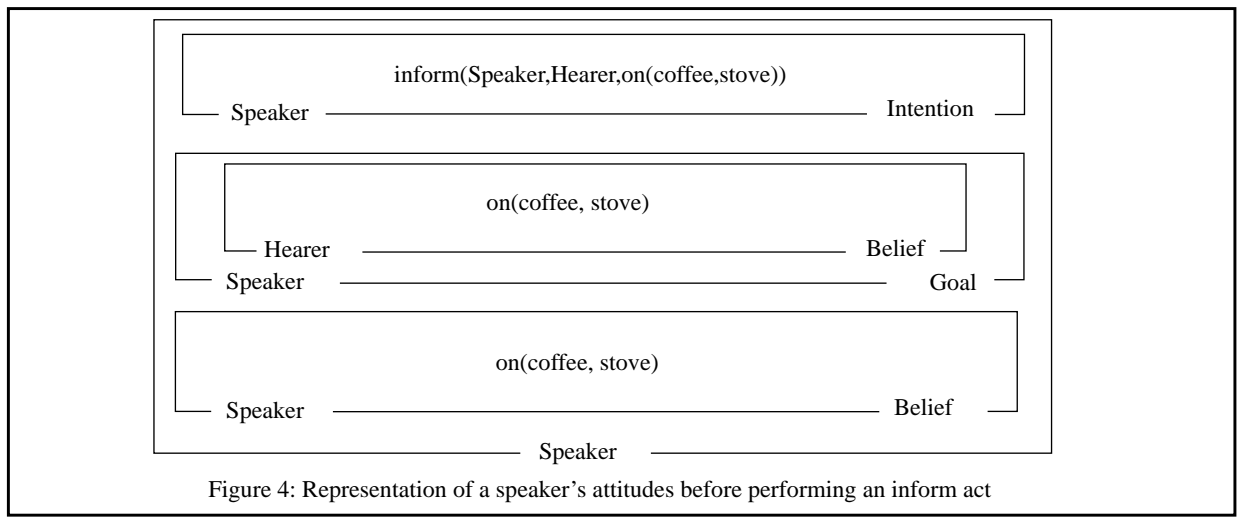

Figure 4: Representation of a speaker's attitudes before performing an inform act

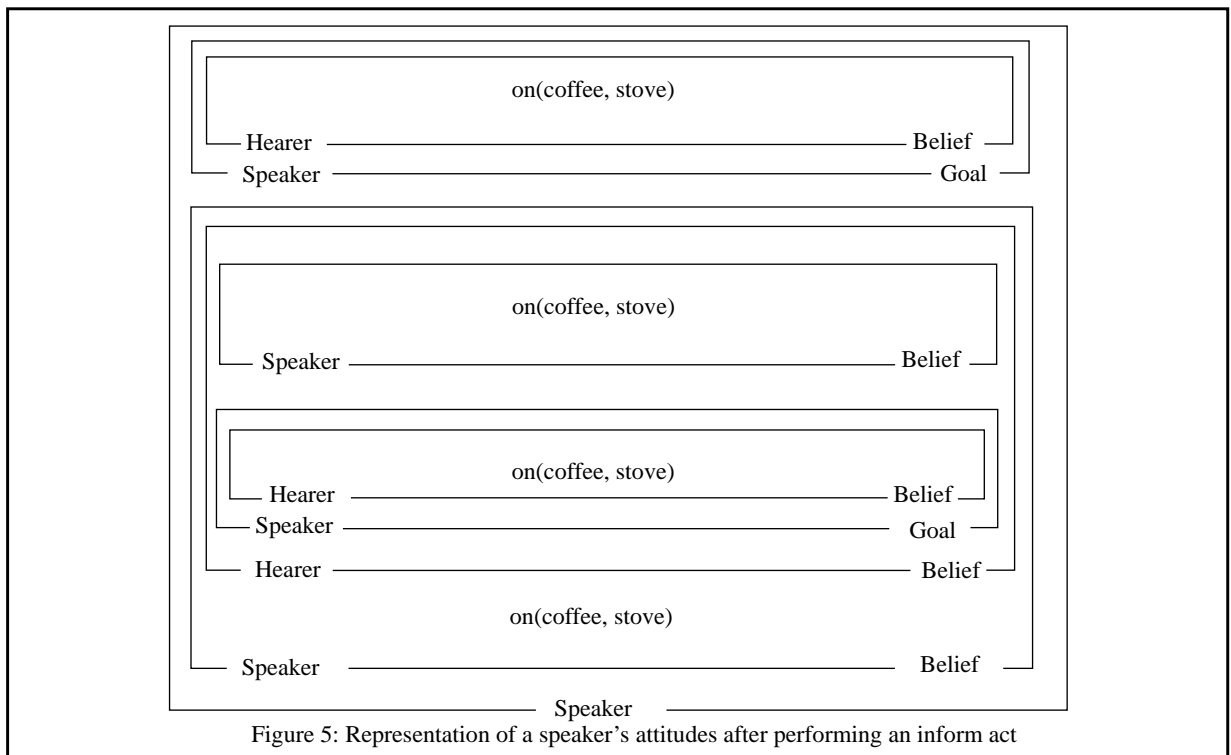
Figure 5: Representation of a speaker's attitudes after performing an inform act

act. Mental acts are detailed in the next section. The update rule for the hearer is the converse of the speaker's:

Update on the Hearer's belief set
*For every condition C in dialogue act performed:*
*default_ascribe(Hearer,Speaker, C)*

That is, given that the speaker has performed an inform act, the hearer can ascribe to the speaker the preconditions of the inform act assuming that the speaker is being cooperative. The effects of the inform act are shown in Figure 6. Note that the hearer's update rule is one level less nested: the preconditions rather than beliefs about the preconditions are ascribed.

### 4.3 Planning and plan simulation in nested belief environments

ViewGen uses a non-linear POCL planner (McAllester and Rosenblatt, 1991) to plan actions to achieve goals. Such a planner is provably correct and complete so that it is guaranteed to find a solution if one exists and only generates valid solutions.

Since ViewGen represents the attitudes of agents in nested environments, it is able to use the planner to simulate other agent's planning. This simulation can be applied to any depth of nested belief e.g. ViewGen can simulate John simulating Mary generating a plan to achieve a given goal by considering its beliefs of John's beliefs of Mary's beliefs, goals and intentions.

Which plan is constructed depends on what the nested agent is believed to believe. Therefore, during nested planning, ViewGen has to reason about which beliefs are held to be true at that level of nesting. However, as mentioned above, belief ascription only is performed as required: we cannot predict which beliefs will be relevant to a plan before the plan is constructed and therefore, ascription must be performed as the plan is generated. To achieve this, both types

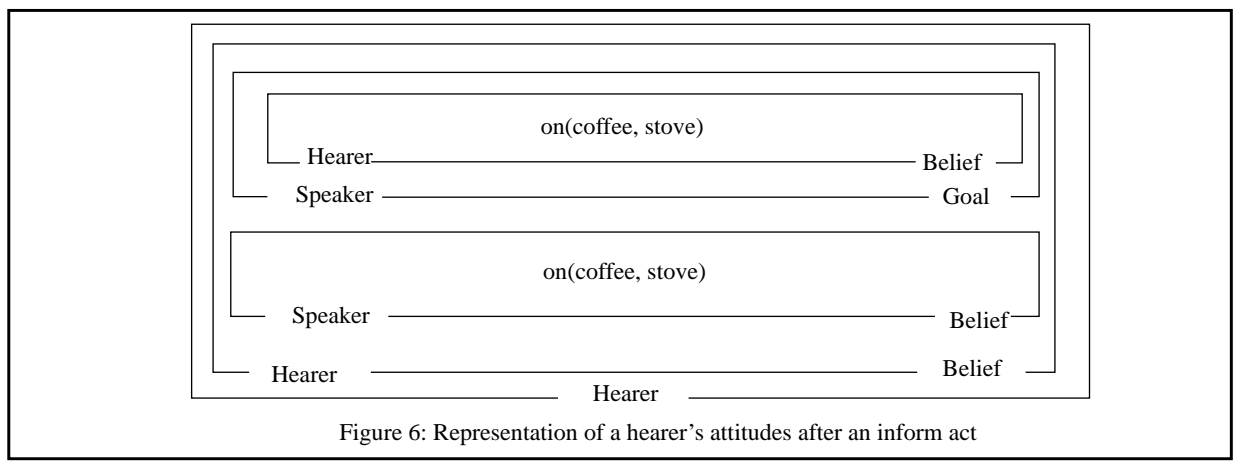
Figure 6: Representation of a hearer's attitudes after an inform act

of ascription are represented in plan operator notation as mental acts. For example, default belief ascription as detailed in section 3.2.1 is represented as:

Default_belief_ascription(Agent1, Agent2, Proposition)
Preconditions: belief(Agent1, Proposition)
belief(Agent1, not(belief(Agent2, not(Proposition))))
Effects: belief(Agent1, belief(Agent2, Proposition))

In addition to pushing outer nested beliefs into inner environments, we require a method of adopting other agent's beliefs by pushing inner nested beliefs into outer environments. For this, we have an accept-belief operator:

Accept_belief(Agent1, Agent2, Proposition)
Preconditions: belief(Agent1, belief(Agent2, Proposition))
not(belief(Agent1, not(Proposition)))
belief(Agent1,trustworthy(Agent2))
Effects: belief(Agent1,Proposition)

That is, if an Agent2 has some belief and Agent1 doesn't hold a contrary belief and believes that Agent2 is trustworthy, then it is acceptable for Agent1 to also believe Agent2's belief. This plays a role in informing where a hearer must decide whether or not to believe the communicated proposition.

During planning, plans are constructed based on the beliefs, goals and intentions which are explicitly present at that level of nesting. However, if a proposition isn't represented at this level of nesting, then the POCL planner must plan ascription actions to determine whether the simulated agent holds the relevant attitude. Therefore, simulated planning involves two types of planning: planning by the agent simulated and planning by ViewGen itself to maintain its belief representation of the agent.

In addition to plan simulation, we have extended the basic POCL algorithm to allow other agent's plans to be recognised. This involves inferring from an agent's performed action, the agent's set of goals he or she is trying to achieve and the plan he or she intends to follow to achieve these goals. This is achieved by collecting together the ascribable goals at the particular level of nesting and attempting to find a plan which achieves at least one of the ascribable goals. Once a plan is generated, any goals achieved by the plan are ascribed.

In both simulation and recognition, once an acceptable plan is generated, the actions and goals in the plan are ascribed to the agent at that level of nesting.

## 5. Conclusions and future work

We have argued that the computation of highly nested belief structures during the performance or recognition of a speech act is implausible. In particular, the concept of mutual belief seems too strong. Instead, we have put forward a theory of speech acts where only the minimal set of beliefs is ascribed at the time of the utterance. If further belief nestings are required then they can be derived using belief ascription techniques as required.

We believe that, for the most part, during normal dialogue, the minimal effects of any speech act are all that are required. However, our approach allows highly nested belief structures to be computed on demand if required, for example, to understand non-conventional language use.

Future work includes the attachment of a robust dialogue parser. We also intend to link ViewGen to the LaSie information extraction platform (Gaizaukas et al, 1995) so as to develop a testable belief set empirically derived from a small medical domain corpus.